\documentclass[10pt,twocolumn,letterpaper]{article}

\usepackage[pagenumbers]{cvpr} 

\definecolor{cvprblue}{rgb}{0.21,0.49,0.74}
\usepackage[pagebackref,breaklinks,colorlinks,allcolors=cvprblue]{hyperref}
\usepackage{algorithm}
\usepackage{algpseudocode}
\usepackage{xcolor}
\usepackage{multirow}
\usepackage{adjustbox}
\usepackage[most]{tcolorbox}
\usepackage{colortbl}
\usepackage{amsmath}
\usepackage{marvosym}
\usepackage{gradient-text}

\usepackage[dvipsnames]{xcolor}

\definecolor{myblue}{HTML}{4E95D9}
\definecolor{bluedots}{HTML}{215F9A}
\definecolor{redtriangle}{HTML}{B02318}
\definecolor{greendots}{HTML}{56835D}

\newcommand{\legendbox}[2]{%
  \begingroup
  \setlength{\fboxsep}{0pt}%
  \setlength{\fboxrule}{0.5pt}%
  \fcolorbox{#2}{#1}{%
    \rule{0pt}{1.4ex}\rule{1.2em}{0pt}
  }%
  \endgroup
}

\newcommand{\bluepoint}{\textcolor{bluedots}{\LARGE$\bullet$}\ }
\newcommand{\greenpoint}{\textcolor{greendots}{\LARGE$\bullet$}\ }
\newcommand{\redtriangle}{\raisebox{0.05ex}{\textcolor{redtriangle}{\large$\blacktriangle$}}}
\newcommand{\titlecolor}[1]{\gradientRGB{#1}{0,0,139}{0,191,255}}

\usepackage{amssymb}
\usepackage{pifont}
\newcommand{\cmark}{\ding{51}}%
\newcommand{\xmark}{\ding{55}}%

\def\confName{CVPR}
\def\confYear{2026}

\title{\titlecolor{VideoSeek}: Long-Horizon Video Agent with Tool-Guided Seeking}

\author{%
   Jingyang Lin\textsuperscript{1,2}\footnotemark[1]\hspace{0.28cm} Jialian Wu\textsuperscript{1~\textrm{\Letter}}\hspace{0.28cm} Jiang Liu\textsuperscript{1} \hspace{0.28cm} Ximeng Sun\textsuperscript{1} \hspace{0.28cm} Ze Wang\textsuperscript{1} \hspace{0.28cm} Xiaodong Yu\textsuperscript{1}\\Jiebo Luo\textsuperscript{2} \hspace{0.28cm} Zicheng Liu\textsuperscript{1}\hspace{0.28cm}Emad Barsoum\textsuperscript{1} \\
    \textsuperscript{1}AMD\hspace{0.28cm}\textsuperscript{2}University of Rochester \\
   \small \textbf{Code:} \href{https://github.com/jylins/videoseek}{\texttt{\textbf{github.com/jylins/videoseek}}}
 }

\begin{document}

\twocolumn[{%
    \renewcommand\twocolumn[1][]{#1}%
    \maketitle%
    \vspace{-6mm}
\centering \centering
\includegraphics[width=0.62\textwidth]{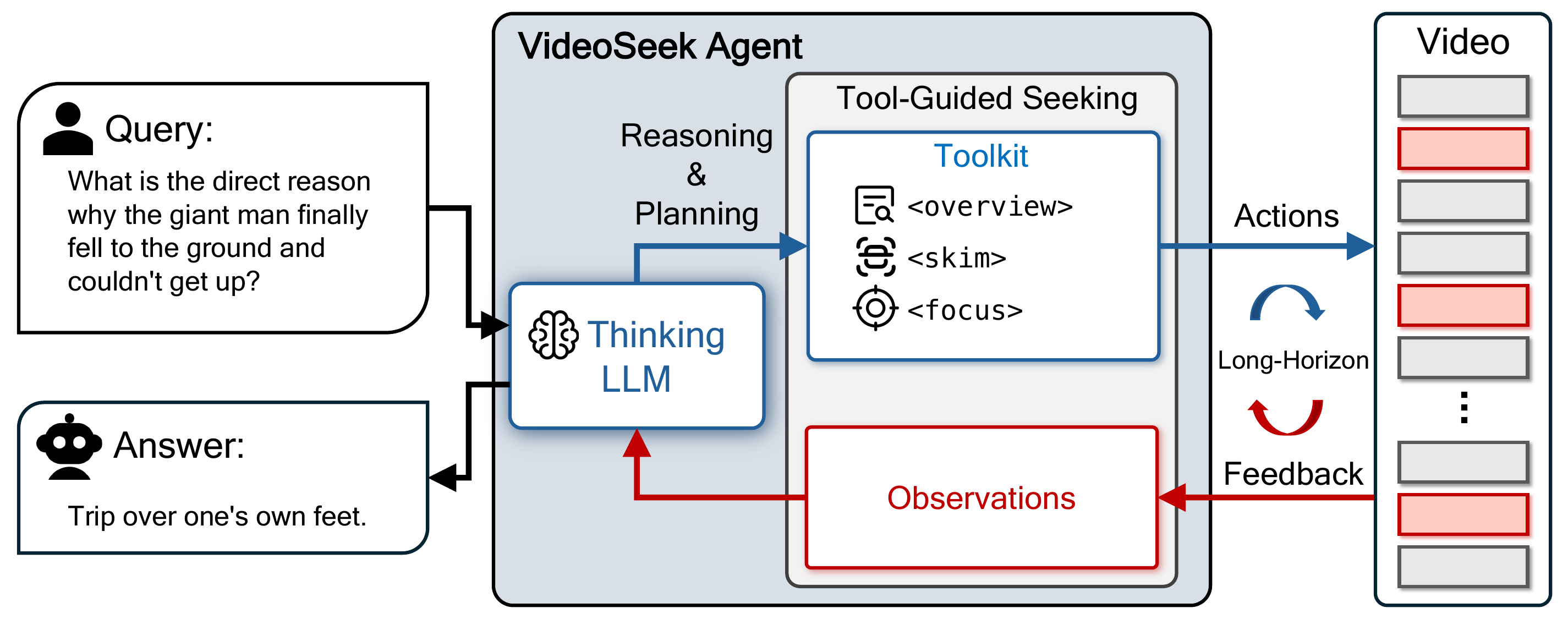}\hfill\includegraphics[width=0.34\textwidth]{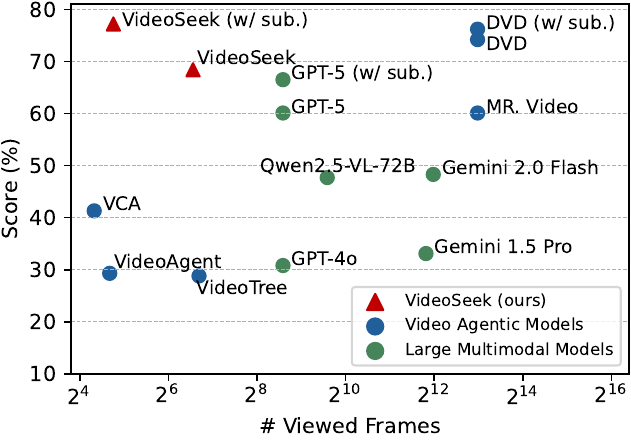}
\captionsetup[figure]{}\vspace{-3mm}
\captionof{figure}
{
\textbf{Overview of VideoSeek}.
\emph{Left}: VideoSeek is a long-horizon video agent that actively seeks answer-critical evidence, guided by video logic flow. Given a query and a video, a thinking LLM reasons over accumulated observations, plans the next step, and selects a tool from the toolkit. The selected tool gathers new evidence from the video ({\legendbox{red!20}{red}: viewed frames; \legendbox{gray!20}{gray}}: unseen frames), which is fed back to the thinking LLM in a think–act–observe loop until sufficient evidence is collected to produce the final answer.
\emph{Right}: Accuracy \emph{vs.} number of viewed frames on LVBench~\cite{wang2025lvbench}. \bluepoint denote video agentic models and \greenpoint denote standalone LMMs. \redtriangle VideoSeek (\emph{w/ subtitles}) achieves the best performance while processing only about $1/300$ as many frames as the second-best video agent.
}\vspace{4.2mm}
\label{fig:overview}
}]

\begin{abstract}
Video agentic models have advanced challenging video-language tasks.
However, most agentic approaches still heavily rely on greedy parsing over densely sampled video frames, resulting in high computational cost.
We present VideoSeek, a long-horizon video agent that leverages video logic flow to actively seek answer-critical evidence instead of exhaustively parsing the full video.
This insight allows the model to use far fewer frames while maintaining, or even improving, its video understanding capability.
VideoSeek operates in a think–act–observe loop with a well-designed toolkit for collecting multi-granular video observations.
This design enables query-aware exploration over accumulated observations and supports practical video understanding and reasoning.
Experiments on four challenging video understanding and reasoning benchmarks demonstrate that VideoSeek achieves strong accuracy while using far fewer frames than prior video agents and standalone LMMs.
Notably, VideoSeek achieves a 10.2 absolute points improvement on LVBench over its base model, GPT-5, while using 93\% fewer frames.
Further analysis highlights the significance of leveraging video logic flow, strong reasoning capability, and the complementary roles of toolkit design.
\end{abstract}
\section{Introduction}\label{sec:intro}
\begingroup
\renewcommand{\thefootnote}{\fnsymbol{footnote}}
\footnotetext[1]{Work was done during the internship at AMD. $^{\textrm{\Letter}}$ Corresponding author: \href{mailto:Jialian.Wu@amd.com}{Jialian.Wu@amd.com}.}
\endgroup
\setcounter{footnote}{0}
Video-language understanding~\cite{lazarus1973multimodal,mcgurk1976hearing,caba2015activitynet,xu2016msr} requires perceiving and reasoning over the video streams and the natural-language instructions to interpret user intent.
Its practical impact spans a wide range of applications, including multimodal assistants~\cite{qian2025dispider,weerasinghe2024real}, autonomous driving~\cite{jiang2025survey,tian2024drivevlm}, and vision-guided robotics~\cite{zitkovich2023rt,song2025robospatial}.
Recent advancements in large language models (LLMs)~\cite{brown2020language,ouyang2022training,bai2023qwen,touvron2023llama} and large multimodal models (LMMs)~\cite{li2024llava,liu2023visual,wang2024qwen2} have successfully pushed the limits of video-language understanding, encouraging a surge of Video-LMMs~\cite{lin2024video,zohar2025apollo,lin2025unleashing,zhang2025video,shu2025video,tang2025video,xu2025slowfast} that achieve promising performance on standard video-language tasks~\cite{chen2011collecting,song2015tvsum,xu2016msr,krishna2017dense,lin2023videoxum,sanders-etal-2024-grounding}.
However, these methods mostly follow a single-pass paradigm, which is often insufficient for more challenging settings, such as long-form video understanding~\cite{wang2025lvbench,wu2024longvideobench,chandrasegaran2024hourvideo,fu2025video} and complex video reasoning~\cite{cheng2025video,fu2025video}.
Attributed to the development of agentic LLMs with stronger reasoning capabilities~\cite{wei2022chain,yao2022react,wang2023plan,shinn2023reflexion,lin-etal-2025-facilitating}, recent video agentic approaches~\cite{zhang2025deep,ma2025drvideo,wang2024videoagent,wang2025videotree} treat video understanding as a long-horizon task~\cite{wei2025browsecomp,li2025mm,geng2025webwatcher}, which demand a long sequence of reasoning, planning, and evidence gathering.

Despite advancements in video agentic models, most existing video agents~\cite{ma2025drvideo,zhang2025deep,pang2025mr} still rely on heavy and expensive video preprocessing.
In particular, they densely parse videos at nontrivial frame rates (\eg, 0.2 - 2 FPS) and translate visual content into detailed textual descriptions or structured memories. For example, DrVideo~\cite{ma2025drvideo} converts a long video into a long
document at 0.2 FPS, while DVD agent~\cite{zhang2025deep} and MR. Video~\cite{pang2025mr} build multi-granular video descriptions at 2 FPS.
Although such preprocessing can improve accuracy, its cost scales poorly with video length, especially for long videos.
More importantly, this heavy preprocessing is often unnecessary: on LVBench~\cite{wang2025lvbench}, over 80\% of questions can be answered by inspecting less than 5\% of the original video.
It suggests that exhaustively annotating multi-granular information is inefficient and unnecessary. Instead of exhaustive parsing, this work therefore explores a more efficient agentic paradigm that solves complex video-language tasks.

Humans rarely solve video QA by watching every frame from beginning to end.
Instead, they typically use the video’s temporal and causal structure~\cite{zhou2018temporal,CLEVRER2020ICLR} to infer where useful evidence is likely to appear, quickly build a rough storyline, inspect promising intervals, and zoom in only when fine-grained details are needed.
This observation motivates a different agentic paradigm: rather than greedily parsing the full video, a model should actively seek informative evidence by leveraging video logic flow.

Motivated by this intuition, we propose \textbf{VideoSeek}, a long-horizon video agent that leverages the video logic flow to actively seek answer-critical evidence throughout the video, as shown in Figure~\ref{fig:overview} (\emph{left}).
Concretely, VideoSeek follows a \emph{think–act–observe} loop~\cite{yao2022react}.
At each step, the agent reasons over the query and accumulated observations, plans the next action to invoke an appropriate tool, and incorporates the returned observations into subsequent reasoning.
To support efficient navigation, we design a light-weight multi-granular toolkit consisting of three tools:
(i) \texttt{<overview>} rapidly scans the video to form a coarse storyline;
(ii) \texttt{<skim>} probes candidate intervals at low cost;
(iii) \texttt{<focus>} closely inspects short clips for answer-critical details.
Together, these tools enable VideoSeek to watch a video at multiple granularities, exhibiting human-like seeking and reasoning behaviors.

Compared to the prior video agents, a key innovation of this work is that VideoSeek executes \emph{long-horizon reasoning and exploration} over the accumulating observations in the \emph{full} conversation history, rather than relying on a prebuilt video database~\cite{ma2025drvideo,zhang2025deep} or a carefully maintained memory buffer~\cite{yang2025vca}.
Based on the evolving observations, this design adaptively adjusts tool-calling strategies to make exploration more flexible, more targeted, and more efficient. As a result, VideoSeek processes far fewer frames while maintaining, and even improving performance.

Empirically, we evaluate the proposed VideoSeek agent on four challenging benchmarks spanning both \textbf{long-form video understanding} and \textbf{complex video reasoning}, including LVBench~\cite{wang2025lvbench}, Video-MME~\cite{fu2025video}, LongVideoBench~\cite{wu2024longvideobench}, and Video-Holmes~\cite{cheng2025video}. VideoSeek consistently achieves strong accuracy under a sparse visual budget. As shown in Figure~\ref{fig:overview} (\emph{right}), VideoSeek achieves the best score while using far fewer frames than other strong peer video agents.
Furthermore, we conduct comprehensive analysis, which highlights the importance of leveraging logic flow, strong reasoning capability, and comprehensive toolkit design for video agentic models.

To summarize, our main contributions are three-fold: 
\begin{itemize}
    \item We propose the VideoSeek, a long-horizon video agent that actively seeks informative evidence by exploiting video logic flow, instead of exhaustively parsing densely sampled frames.
    \item Extensive experiments on long-form video understanding and complex reasoning benchmarks demonstrate that VideoSeek achieves state-of-the-art performance while using far fewer frames than prior video agents, highlighting its efficiency and applicability.
    \item Our comprehensive analysis underscores the importance of leveraging video logic flows, strong reasoning capability, and a well-designed toolkit in developing effective and efficient video agentic models.
\end{itemize}

\section{Related Work}
\label{sec:related_work}

\begin{figure*}
    \includegraphics[width=1\textwidth]{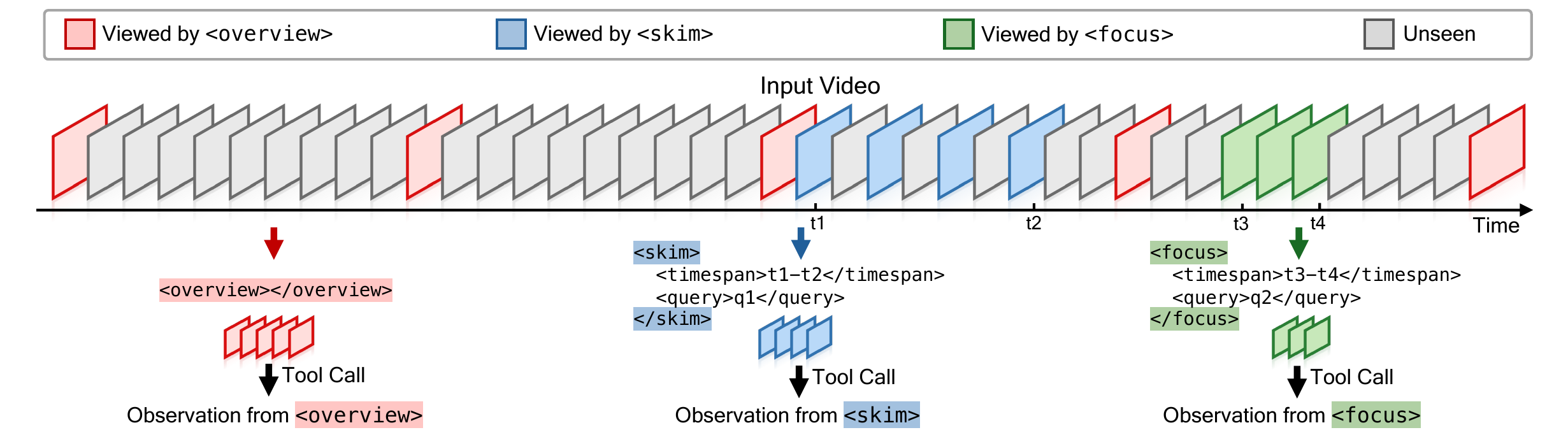}\vspace{0mm}
    \caption{Toolkit of the VideoSeek agent, including \texttt{<overview>}, \texttt{<skim>}, and \texttt{<focus>} tools. \emph{Left}: \texttt{<overview>} rapidly scans the entire video to build a coarse storyline and highlight promising intervals. \emph{Middle}: \texttt{<skim>} takes a quick glance at these candidate intervals (\ie, $t_1$ to $t_2$) at low cost to check whether query-relevant evidence is nearby. \emph{Right}: \texttt{<focus>} zooms in on a fine-grained clip (\ie, $t_3$ to $t_4$) with dense inspection to obtain answer-critical observations. Red, blue, and green boxes denote frames viewed by \texttt{<overview>}, \texttt{<skim>}, and \texttt{<focus>}, respectively, while gray boxes indicate unseen frames.}
    \label{fig:toolkit}
\end{figure*}

\textbf{Video-Language Models}.
The rapid advances of LLMs~\cite{brown2020language,ouyang2022training,bai2023qwen,touvron2023llama} and LMMs~\cite{li2024llava,liu2023visual,wang2024qwen2} have substantially accelerated progress in video-language models, particularly video large multimodal models.
Early attempts~\cite{damonlpsg2024videollama2,damonlpsg2023videollama,xu2024pllava} extend image-language architectures to videos through video-specific adapters between the vision encoder and language decoder.
Inspired by the success of synthetic instruction-following data in the image-language domain~\cite{liu2023visual}, subsequent work then shifts its attention to constructing high-quality synthetic video instruction-following data. Follow-up works~\cite{li2023videochat,grunde2021agqa} adopt template-based QA generation on existing video captioning datasets.
By leveraging powerful LMMs~\cite{gpt4v2023,gpt4o2024}, recent works~\cite{chen2024sharegpt4video,zhang2025video} produce high-quality video captions and diverse QA pairs.
Apollo~\cite{zohar2025apollo} further highlights the significance of text, image, and video data mixtures during training.
As video-language models begin to saturate performance on basic video-language tasks~\cite{chen2011collecting,xu2016msr,krishna2017dense}, long-form video-language understanding~\cite{wang2025lvbench,chandrasegaran2024hourvideo,xu2025slowfast,lin2025unleashing,shen2024longvu,shu2025video,xue2024longvila} has attracted increasing attention, which demands parsing hour-scale videos while maintaining token efficiency. Meanwhile, the strong reasoning capabilities of recent LLMs~\cite{guo2025deepseek,openai2025o3} have motivated exploration of complex video reasoning~\cite{cheng2025video,feng2025video}.
However, these methods still largely follow a single-pass paradigm, in which a fixed set of frames is processed before directly producing an answer. Such a formulation is often insufficient for challenging scenarios that require iterative evidence gathering and long-horizon reasoning.
Beyond single-pass paradigm, this work regards video-language tasks as long-horizon problems requiring iterative planning, targeted evidence gathering, and continual evaluation of whether the collected evidence is sufficient to answer.

\noindent \textbf{Video Agentic Models}.
Early video agentic approaches rely on manually designed and human-crafted workflows.
VideoAgent~\cite{wang2024videoagent} pioneers using an LLM as a central agent that iteratively inspects key video frames and then retrieves query-relevant frames via CLIP~\cite{radford2021learning}. 
Subsequent works~\cite{wang2025videotree,yang2025vca} refine this idea by performing a coarse-to-fine, tree-structured search over video segments to identify informative frames.
Beyond pure search, later studies~\cite{ma2025drvideo,luo2025video,pang2025mr} construct a comprehensive video database for query-relevant information retrieval.
Instead of relying on predefined workflows, recent studies~\cite{zhang2025deep,tian2025ego} develop autonomous and adaptive agentic paradigms with tool use for diverse and real-world scenarios.
Built on search-centric toolkits and multi-granular video databases, DVD~\cite{zhang2025deep} proactively discovers and extracts crucial evidence from the given video.
Ego-R1 Agent~\cite{tian2025ego} proposes chain-of-tool-thought reasoning to iteratively decompose complex video reasoning tasks and invoke specialized tools to resolve them.
However, most video agentic approaches either depend on a prebuilt video database~\cite{ma2025drvideo,luo2025video,pang2025mr,zhang2025deep,tian2025ego} or greedily scan the entire video~\cite{wang2024videoagent,wang2025videotree}.
Although such paradigms can capture detailed video content, their expensive preprocessing cost limits practicality in real-world settings and hinders scaling to long-form videos.
In contrast, VideoSeek leverages the inherent logic flows within videos and actively seeks informative frames based on the accumulating observations over the long-horizon conversation history, thereby avoiding densely parsing the full video.
\section{Methodology}

\subsection{Problem Formulation}
Conventional video-language tasks require a model to generate an answer $\mathbf{Y}$ given a query $\mathbf{Q}$ and a video $\mathbf{X}$, by modeling the conditional probability:
\begin{equation}
    p(\mathbf{Y} \mid \mathbf{X}, \mathbf{Q}).
\end{equation}
In this work, we instead treat video-language tasks as \textbf{long-horizon problems}, where the model iteratively \emph{think}, \emph{act}, and \emph{observe} before producing a final answer.
At each reasoning step $t$, the video agent produces a \emph{think–act–observe} triplet $\langle z_t, a_t, o_t \rangle$, where $z_t$ denotes the internal reasoning trace, $a_t$ indicates the selected action (\ie, a specific tool calling), and $o_t$ refers to the resulting observation.
Over $n$ reasoning steps, these triplets forms a trajectory $\tau$:
\begin{equation}
    \tau = \big(\langle z_1, a_1, o_1 \rangle, \dots, \langle z_t, a_t, o_t \rangle, \dots, \langle z_n, a_n, o_n \rangle\big),
\end{equation}
where $n$ is the total number of reasoning turns used for the given query.
From this long-horizon perspective, solving a video-language task amounts to predicting both the full reasoning trajectory $\tau$ and the final answer $\mathbf{Y}$ conditioned on the video $\mathbf{X}$ and query $\mathbf{Q}$:
\begin{equation}
    p(\tau, \mathbf{Y} \mid \mathbf{X}, \mathbf{Q}).
\end{equation}
Intuitively, the agent first explores the video and builds a trajectory $\tau$, and then uses the accumulated evidence to generate the final answer.
This process can be factorized as:
\begin{equation}\label{eq4}
    p(\tau, \mathbf{Y} \mid \mathbf{X}, \mathbf{Q})
    = p(\tau \vert \mathbf{X}, \mathbf{Q}) \cdot p(\mathbf{Y} \mid \mathbf{X}, \mathbf{Q}, \tau),
\end{equation}
where $p(\tau \mid \mathbf{X}, \mathbf{Q})$ captures long-horizon reasoning and evidence seeking, and $p(\mathbf{Y} \mid \mathbf{X}, \mathbf{Q}, \tau)$ models answer generation conditioned on the accumulated trajectory.

\subsection{Toolkit Design}
To support efficient long-horizon reasoning under a limited visual budget, VideoSeek is equipped with a lightweight but effective toolkit of three specialized video-analytic tools, as shown in Figure~\ref{fig:toolkit}. Inspired by human behaviors, each tool operates at a different temporal granularity of the video, allowing the agent to strategically trade off between global coverage and fine-grained details. The agent invokes these tools on demand during reasoning, progressively narrowing the search space from the full video to answer-critical clips.

\noindent \textbf{Overview Tool}.
The \texttt{<overview>} tool provides a coarse global summary of the full video to establish a brief storyline for subsequent exploration.
It uniformly samples a fixed number of frames across the entire timeline and produces brief descriptions.
This summary gives the agent an initial map of the video structure (\eg, storyline, key characters, and locations) without exhaustively watching the full video.
Such global information is crucial for long-horizon reasoning, as it helps the agent form an initial plan and pinpoint query-relevant regions, while keeping the observation cost low.
The overview is primarily used at the beginning to identify promising regions for further exploration.

\noindent \textbf{Skim Tool}.
The \texttt{<skim>} tool performs a coarse-grained scan of a candidate segment that is still too long for dense analysis.
This tool provides a coarse-grained scan of a selected video segment that is still too long for frame-by-frame inspection.
Once the agent has inferred a broad candidate region of interest, it can call the \texttt{<skim>} tool on this segment.
Given a selected interval, the tool uniformly samples a small number of frames and highlights those most relevant to the current query.
Instead of inspecting every frame, this step significantly narrows down the search space.
In this way, the \texttt{<skim>} tool supports long-horizon reasoning by zooming in gradually: it bridges the gap between a global overview and fine-grained frame-level inspection, helping the agent decide which timespans of a long segment deserve deeper analysis.
The agent may invoke \texttt{<skim>} multiple times on different candidate segments to progressively narrow the search space.

\noindent \textbf{Focus Tool}.
The \texttt{<focus>} tool enables fine-grained analysis of a short video clip at a higher frame rate and serves as the final``close-up'' examination step.
When the agent needs to verify or extract precise information that coarser tools cannot provide, it invokes the \texttt{<focus>} tool on a specific temporal interval at a high frame rate (\ie, 1 FPS), along with an agent-formulated query.
By operating at this level, the agent can capture subtle details, such as reading text on a sign, recognizing a character’s face, counting objects, or confirming an action that occurs within a brief moment.
As a result, the \texttt{<focus>} tool plays a key role in ensuring the accuracy of the final answer and in preventing errors introduced by coarse-grained tools.

\begin{algorithm}[t]
\caption{VideoSeek Agent Workflow}
\label{alg:videoseek}
\begin{algorithmic}[1]
\Require User query $\mathbf{Q}$; video $\mathbf{X}$; system instruction $\mathcal{I}$; thinking model $\theta_{\text{think}}$; toolkit $\mathcal{T}$; max turn limit $N$. 
\Ensure Answer $\mathbf{Y}$.

\State Initialize reasoning trajectory $\tau \gets \langle \mathcal{I}, \mathbf{Q} \rangle$
\State Initialize toolkit $\mathcal{T} \gets \mathcal{T} \cup \texttt{<answer>}$
\State $\mathbf{Y} \gets \varnothing$
\For{$t = 1$ to $N$}
    \State $(z_t, \alpha_t) \gets \theta_{\text{think}}(\tau)$ \Comment{reasoning and tool-planning}
    \If{$|\alpha_t| = 1$ \textbf{ and } $\alpha_t = \texttt{<answer>}$}
        \State $\mathbf{Y} \gets \textsc{ParseAnswer}(\alpha_t)$
        \State \textbf{break}
    \EndIf
    \State $o_t \gets \textsc{CallTools}(\alpha_t, \mathbf{X}, \mathcal{T})$
    \State $\tau \gets \tau \ \cup \langle z_t, \alpha_t, o_t \rangle$ \Comment{append to trajectory}
\EndFor
\If{$\mathbf{Y} = \varnothing$}
    \State $\tau \gets \tau \, \cup \mathcal{I}_{\text{answer}}$ \Comment{add direct-answer instruction}
    \State $\mathbf{Y} \gets \theta_{\text{think}}(\tau)$
\EndIf
\State \Return $\mathbf{Y}$
\end{algorithmic}
\end{algorithm}

\subsection{Agentic Workflow of VideoSeek}
Building on the above toolkit, VideoSeek operates in a ReAct-style~\cite{yao2022react} workflow, where reasoning and tool use are interleaved. As illustrated in Figure~\ref{fig:overview} (\emph{left}), the core of this workflow is a loop in which the agent iteratively thinks, acts, and observes, progressively gathering and accumulating evidence until it is confident enough to answer a given query. At each turn, the VideoSeek agent executes the \emph{think-act-observe} workflow:
\begin{itemize}
    \item \textbf{Thought}. The thinking LLM first reasons over the user query and the current trajectory, assessing what observations has already been collected, what remains uncertain, and whether the existing evidence is sufficient to answer the question.
    \item \textbf{Action}. Based on the above reasoning, the agent either decides to invoke an \texttt{<answer>} action or selects one of video tools with an appropriate interval and query to gather the most informative next piece of evidence.
    \item \textbf{Observation}. The selected tool returns new evidence, which is appended to the trajectory and used in the next round of reasoning.
\end{itemize}
The resulting \emph{think-act-observe} triplet is then added to the trajectory and used in the next turn. If uncertainty remains, the agent actively seeks clues via tool using, repeating until the agent decides that sufficient evidence has been gathered or the maximum turn limit is reached.

Formally, we elaborate the agentic workflow of VideoSeek in Algorithm~\ref{alg:videoseek}.
Given a user query $\mathbf{Q}$ and its corresponding video $\mathbf{X}$, the agent maintains a reasoning trajectory $\tau$, initialized with the system instruction $\mathcal{I}$ and the user query $\mathbf{Q}$.
The agent is powered by a thinking model $\theta_{\text{think}}$ and a toolkit $\mathcal{T}$ that consists of three multi-granular view tools (\texttt{<overview>}, \texttt{<skim>}, \texttt{<focus>}) and the \texttt{<answer>} tool.
At each turn $t$, the thinking model reads the previous trajectory $\tau$ (including all past thoughts, actions, and observations) and outputs a reasoning trace $z_t$ together with a concrete tool plan $\alpha_t$.
If $\alpha_t$ contains only a single \texttt{<answer>} call, the agent parses this output with \textsc{ParseAnswer}($\cdot$) and stops. Otherwise, the agent executes the planned tools on the video via \textsc{CallTools}($\cdot$), obtaining the corresponding observations $o_t$ from $\mathbf{X}$. The new triplet $\langle z_t, \alpha_t, o_t \rangle$ is then appended to $\tau$ and becomes part of the context for the next turn.
If no answer is produced within the turn limit $N$, we add a direct-answer instruction $\mathcal{I}_{\text{answer}}$ to the trajectory and invoke $\theta_{\text{think}}$ once more to synthesize the final answer $\mathbf{Y}$ from the accumulated evidence.

Humans rarely inspect every single frame to understand a video. Instead, they quickly form a rough understanding of the storyline, then jump to segments where the answer is likely to appear, and only re-watch short clips carefully when details matter.
VideoSeek explicitly mirrors this pattern through its \emph{think–act–observe} loop: each new observation is used to refine the agent’s belief about where the answer might lie, which in turn guides subsequent tool calls.
As a result, the model can process substantially fewer frames while maintaining, and even surpassing, the video-language comprehension capability of dense-parsing baselines (See Section~\ref{sec:exp}).
In our formulation in Eq.~(\ref{eq4}), this advantage appears as a more efficient, query-aware estimation of $p(\tau \mid \mathbf{X}, \mathbf{Q})$, where the trajectory focuses on a few highly informative observations rather than exhaustive coverage.
This compact yet informative trajectory, in turn, makes it easier for $p(\mathbf{Y} \mid \mathbf{X}, \mathbf{Q}, \tau)$ to generate accurate answers.
\section{Experiments}\label{sec:exp}

\subsection{Experimental Setting}
\noindent \textbf{Evaluation Benchmarks}. 
We evaluate VideoSeek on four video-language benchmarks spanning both long-form video understanding and complex video reasoning:
\begin{itemize}
    \item \textbf{LongVideoBench}~\cite{wu2024longvideobench} contains web videos of varying lengths up to one hour, together with subtitles, covering diverse themes and evaluating detailed retrieval and reasoning over long videos. We report results on its \emph{long} split, which comprises 564 questions from 188 videos with durations between 900 and 3600 seconds.
    \item \textbf{Video-MME}~\cite{fu2025video} is a comprehensive multimodal benchmark for long video understanding across diverse video types and temporal ranges. We evaluate on its \emph{long} subset, which includes 900 questions from 300 videos with an average duration of 2,466 seconds.
    \item \textbf{LVBench}~\cite{wang2025lvbench} focuses on long-term memory and extended comprehension over multimodal inputs, consisting of 1,549 multiple-choice questions constructed from 103 hour-long videos.
    \item \textbf{Video-Holmes}~\cite{cheng2025video} is a complex video reasoning benchmark built from 270 manually annotated suspense short films, containing 1,837 questions across seven tasks that require models to actively locate, connect, and interpret multiple visual clues scattered throughout the video.
\end{itemize}

\begin{table*}[t]
\centering
\caption{Comparison on long-form video benchmarks, including LVBench, VideoMME, and LongVideoBench. \#Frames denotes the number of processed frames. For LVBench and VideoMME, we report results both with and without subtitles. \textbf{Bold} marks the best performance, and \underline{underline} marks the second-best.}\vspace{-2mm}
\label{tab:videoseek_comparison}
\setlength{\tabcolsep}{4pt}
\begin{adjustbox}{width=1\textwidth}\setlength\tabcolsep{5pt}
\begin{tabular}{lcccccccccc}
\toprule
\multirow{2}{*}{\textbf{Method}} &
\multicolumn{2}{c}{\textbf{LVBench} (w/o sub)} &
\multicolumn{2}{c}{\textbf{LVBench} (w/ sub)} &
\multicolumn{2}{c}{\textbf{VideoMME} (w/o sub)} &
\multicolumn{2}{c}{\textbf{VideoMME} (w/ sub)} &
\multicolumn{2}{c}{\textbf{LongVideoBench}} \\
\cmidrule(lr){2-3}  \cmidrule(lr){4-5} \cmidrule(lr){6-7}  \cmidrule(lr){8-9} \cmidrule(lr){10-11}  
& \#Frames & Test & \#Frames & Test & \#Frames & Long & \#Frames & Long & \#Frames & Long (val) \\
\midrule
\multicolumn{11}{>{\columncolor[gray]{.88}}l}{\emph{Large Multimodal Models}} \\
\midrule
Qwen2.5-VL-72B~\cite{bai2025qwen2}        & 768  & 47.3 & -   & -    & 256   & 53.2 & 256   & 64.4 & -    & -    \\
GPT-4o~\cite{gpt4o2024}                & 384  & 30.8 & -   & -    & 384   & 65.3 & 384   & 72.1 & 256  & 60.9 \\
Gemini 1.5 Pro~\cite{gemini15pro2024}       & 3,600 & 33.1 & -   & -    & 1,233  & 67.4 & 1,233  & 77.4 & 256  & 58.6 \\
Gemini 2.0 Flash~\cite{gemini20flash2024}      & 4,037 & 48.3 & -   & -    & 1,233  & 63.0 & -  & -    & 256  & 45.7 \\
GPT-5~\cite{gpt52025} (Base)          & 384  & 60.1 & 384 & 66.5 & 384   & \underline{67.9} & 384   & \underline{78.1} & 384  & 64.5 \\
\midrule
\multicolumn{11}{>{\columncolor[gray]{.88}}l}{\emph{Video Agentic Models}} \\
\midrule
VideoAgent~\cite{wang2024videoagent}            & 25.5  & 29.3 & -   & -    & 24.6  & 46.4 & -     & -    & -    & -    \\
VideoTree~\cite{wang2025videotree}             & 103.2 & 28.8 & -   & -    & 98.0    & 53.1 & -     & -    & -    & -    \\
DrVideo~\cite{ma2025drvideo}               & -     & -    & -   & -    & 493.2 & 51.7 & 493.2 & 71.7 & -    & -    \\
VCA~\cite{yang2025vca}                   & 20.0    & 41.3 & -   & -    & 18.1  & 54.2 & -     & -    & -    & -    \\
MR. Video~\cite{pang2025mr}            & 8,074  & 60.8 & -   & -    & 4,932  & 61.8 & 4,932  & -    & 2,816 & 61.6 \\
DVD~\cite{zhang2025deep}  & 8,074  & \textbf{74.2} & 8,074 & \underline{76.0} & 4,932 & 67.3 & 4,932  & -    & 2,816 & \underline{68.6} \\
\midrule
\rowcolor{myblue!15} \textbf{VideoSeek} (\emph{ours})      & 92.3    & \underline{68.4} & 27.2 & \textbf{76.7} & 60.9 & \textbf{70.1} & 15.9  & \textbf{81.2} & 29.6 & \textbf{73.5} \\
\bottomrule
\end{tabular}
\end{adjustbox}
\end{table*}

\noindent \textbf{Base Models}.
We adopt GPT-5~\cite{gpt52025} as the default thinking LLM in the VideoSeek agent due to its strong \emph{reasoning and tool using} capability. To analyze the role of the underlying reasoning model, we further replace GPT-5 with other alternative LLMs (\eg, o4-mini~\cite{openai2025o3} and GPT-4.1~\cite{openai2025gpt41}) in the ablation study. In addition, we employ GPT-5 to interpret visual content when invoking the view tools in the VideoSeek toolkit.

\noindent \textbf{Implementation Details}.
VideoSeek is a model-agnostic agentic framework, meaning it can be paired with any LMM as the underlying reasoning engine. In this paper, we use GPT-5 as the default LMM within the VideoSeek agent.
The toolkit of VideoSeek consists of three tools, each with task-specific hyperparameters:
\begin{itemize}
    \item \texttt{<overview>} tool uniformly samples $16\alpha$ frames from the entire video to construct a coarse storyline.
    \item \texttt{<skim>} tool operates on relatively long video segments to quickly localize answer-relevant moments. It processes segments of at least $4\alpha$ seconds and samples $4\alpha$ frames.
    \item \texttt{<focus>} tool performs fine-grained analysis on short clips. We set its sampling rate to 1 FPS and cap the clip length at $4\alpha$ seconds.
\end{itemize}
We analyze the effect of $\alpha$ in the Appendix~\ref{sec:tool_frame}. We select the optimal $\alpha$ that achieves best tradeoff between performance and efficiency. Specifically, we set $\alpha=4$ for LVBench and $\alpha=2$ for the other three benchmarks.

For the reasoning trajectory, we set the maximum turn limit $N$ to 20. 
In addition, the official results of GPT-5 on those four video-language benchmarks are not available from public leaderboards or reports; we thereby evaluate GPT-5 on these video-language benchmarks as a reference baseline for our VideoSeek agent. Specifically, we uniformly sample 384 frames per video.

\subsection{Main Results}

\noindent \textbf{Results on Long-form Video Benchmarks}. Table~\ref{tab:videoseek_comparison} reports results on LVBench, Video-MME, and LongVideoBench. Across all three benchmarks, VideoSeek consistently improves over its base model GPT-5 while processing far fewer frames, showing that active evidence seeking is substantially more efficient than dense uniform parsing.
\begin{itemize}
    \item \textbf{LVBench}.
    We evaluate VideoSeek under both subtitle and non-subtitle settings. Without subtitles, VideoSeek achieves 68.4\% accuracy using only 92.3 frames on average, ranking second overall and clearly surpassing all standalone LMMs and most video agents. It improves over its base model GPT-5 by $+8.3$ points while using only $\sim$24\% of its frames, and reaches a performance comparable to the strongest peer agent DVD (74.2\% with 8,074 frames) while consuming only about 1\% of its frame budget. With subtitles, VideoSeek obtains a significant improvement and achieves the best performance among all methods, reaching 76.7\% with fewer frame usage (from 92.3 to 27.2 frames). It not only surpasses GPT-5 with subtitles (66.5\% with 384 frames, $+10.2$ points) but also outperforms the DVD using only 0.3\% of frames. These results demonstrate that VideoSeek can fully exploit subtitle signals while operating under an extremely sparse visual budget.
    \item \textbf{VideoMME} (\emph{long} subset).
    On the VideoMME (\emph{long} subset) without subtitles, VideoSeek achieves 70.1\% accuracy using only 60.9 frames, outperforming all LMMs and video agents. It improves over GPT-5 (67.9\% with 384 frames) by $+2.2$ points while using 84\% fewer of the frames. Compared with Gemini 1.5 Pro (67.4\% with 1,233 frames) and DVD (67.3\% with 4,932 frames), VideoSeek not only yields higher accuracy but also reduces the frame usage to about 1–5\% of these methods. 
    With subtitles, VideoSeek further widens the gap, reaching 81.2\% using only 15.9 frames. This is substantially higher than GPT-5 (78.1\% with 384 frames) and Gemini~1.5~Pro (77.4\% with 1,233 frames), while using around 4\% of GPT-5’s frames and close to 1\% of Gemini’s. Compared to DrVideo (71.7\% with 493.2 frames), VideoSeek gains nearly 10 points with 96\% fewer frame usage.
    \item \textbf{LongVideoBench} (\emph{long} subset).
    On LongVideoBench, VideoSeek again delivers the state-of-the-art performance, achieving 73.5\% accuracy with 29.6 frames on average. This significantly outperforms GPT-5 (64.5\% with 384 frames, $+9.0$ points) while using about 8\% of the frames. Compared with strong peer video agents, VideoSeek surpasses DVD (68.6\% with 2,816 frames) and MR.~Video (61.6\% with 2,816 frames) by $+4.9$ and $+11.9$ points, respectively, using only around 1\% of their frames. These results suggest that VideoSeek is effective and efficient on long-form videos, where the cost of exhaustive parsing becomes extremely expensive.
\end{itemize}
Overall, VideoSeek consistently improves over its base model GPT-5 across all three benchmarks and both with and without subtitles, while reducing the number of processed frames by 76–96\%. 
It demonstrates that actively seeking informative content via tool use enables a more efficient utilization of visual evidence, promoting state-of-the-art performance on long-form video understanding at a low computational cost.

\begin{table*}[t]
\centering
\caption{Comparison on the Video-Holmes. \#Frames denotes the frame usage. Symbol $^\dagger$ indicates that we adopt 1 FPS (default setting of Gemini~1.5~Pro) to estimate the number of viewed frames. \textbf{Bold} marks the best performance, and \underline{underline} marks the second-best. Abbreviations:
SR (Social Reasoning), IMC  (Intention \& Motive Chaining), TCI (Temporal Causal Inference), TA (Timeline Analysis), MHR (Multimodal Hint Reasoning), PAR (Physical Anomaly Reasoning), and CTI (Core Theme Inference).}\vspace{-2mm}
\begin{adjustbox}{width=1\textwidth}\setlength\tabcolsep{12pt}
\begin{tabular}{lcccccccccc}
\toprule
\textbf{Method} & \textbf{\#Frames} & \textbf{SR} & \textbf{IMC} & \textbf{TCI} & \textbf{TA} & \textbf{MHR} & \textbf{PAR} & \textbf{CTI} & \textbf{Overall} \\
\midrule
Qwen2.5-VL-32B~\cite{bai2025qwen2} & 32   & 43.2  & 44.2  & 31.5  & 51.0 & 36.4  & 31.4  & 32.2  & 38.4  \\
SEED-Bench-R1~\cite{chen2025exploring} & 32   & 42.8  & 35.1  & 25.6  & 40.5 & 29.2  & 29.9  & 32.6  & 33.5  \\
VideoChat-R1~\cite{li2025videochat}   & 32   & 42.1  & 38.8  & 24.5  & 39.5 & 29.5  & 27.8  & 29.3  & 33.0  \\
Video-R1~\cite{feng2025video}       & 32   & \underline{48.6}  & 41.7  & 28.9  & 34.5 & 31.0  & 33.5  & 35.9  & 36.5  \\
GPT-4o~\cite{gpt4o2024}         & 32   & 50.0  & \textbf{49.6}  & 38.8  & 30.0 & 44.0  & 39.2  & 37.0  & 42.0  \\
Gemini~1.5~Pro~\cite{gemini15pro2024} & 185.1$^\dagger$   & 52.1  & 48.2  & 34.4  & 26.0 & 39.2  & \textbf{46.4}  & 38.9  & 41.2  \\
Gemini~2.5~Pro~\cite{gemini25pro2025} & 185.1$^\dagger$   & 46.6  & \underline{49.3}  & \textbf{46.9}  & 53.0 & 40.1  & \underline{44.3}  & 37.4  & \underline{45.0}  \\
Gemini~2.0~Flash~\cite{gemini20flash2024} & 185.1$^\dagger$ & 41.8  & 33.7  & 23.1  & 20.5 & 30.1  & 26.8  & 33.7  & 30.6  \\
Gemini~2.0~Flash~Thinking~\cite{gemini20flashthinking2025} & 185.1$^\dagger$ & 43.4 & 46.9 & 43.1 & 51.0 & 37.9 & 43.6 & 39.3 & 43.1 \\
\midrule
GPT-5~\cite{gpt52025} (Base)   & 384  & 47.2 & 43.4  & 40.6 & \underline{53.5} & \underline{46.3}  & 38.1 & \underline{39.6} & 44.1  \\
\rowcolor{myblue!15} VideoSeek (\emph{ours}) & 42.7 & \textbf{56.1} & 43.8 & \underline{45.0} & \textbf{54.5} & \textbf{46.6} & 43.3 & \textbf{41.8} & \textbf{47.3} \\
\bottomrule
\end{tabular}\label{tab:videoholmes}
\end{adjustbox}\vspace{-1mm}
\end{table*}

\begin{table}[t]
\centering
\caption{Comparison of different thinking models used in the VideoSeek agent on LVBench. \#Frames denotes the frame usage. \#Turns indicates the number of \emph{think-act-observe} turns used for obtaining the final answer.}\vspace{-3mm}
\begin{adjustbox}{width=0.49\textwidth}\setlength\tabcolsep{10pt}
\begin{tabular}{lccc}
\toprule
Thinking LLM $\theta_\text{think}$ & \#Frames & \#Turns & LVBench \\
\midrule
\rowcolor{myblue!15}  GPT-5~\cite{gpt52025}    & 92.3  & 4.42 & \textbf{68.4}  \\
\midrule

o4-mini~\cite{openai2025o3}  & 112.6 & 5.08 & \hspace{8mm} 58.5 ({\color{red}-9.9}) \\
GPT-4.1~\cite{openai2025gpt41}  & 74.2  & 2.99 & \hspace{9.7mm} 53.0 ({\color{red}-15.4}) \\
\bottomrule
\end{tabular}\label{tab:think_ablation}
\end{adjustbox}
\end{table}

\noindent \textbf{Results on Complex Video Reasoning Benchmark}.
Table~\ref{tab:videoholmes} presents the evaluation on Video-Holmes~\cite{cheng2025video}, VideoSeek achieves the best overall accuracy of 47.3\% while using only 42.7 frames on average, surpassing strong LMMs and proprietary models such as Gemini~2.5~Pro (45.0\% with 185.1 frames) and its own base model GPT-5 (44.1\% with 384 frames), thus improving performance while cutting frame usage by nearly an order of magnitude. More specifically, VideoSeek excels in several reasoning skills that require integrating long-range narrative evidence: it achieves the highest accuracy on SR (56.1\%), TA (54.5\%), MHR (46.6\%), and CTI (41.8\%), and ranks second on TCI (45.0\%), all while consistently outperforming GPT-5 across almost all dimensions.
Overall, these results demonstrate that even in scenarios demanding complex reasoning, VideoSeek supports effective long-horizon inference under a sparse visual budget.

\subsection{Empirical Analysis}

\noindent \textbf{Effect of Video Logic Flow}.
As shown in Table~\ref{tab:videoseek_comparison}, once subtitles are involved, VideoSeek obtains substantial performance gains on both LVBench and VideoMME while its frame usage drops dramatically. This trend suggests that subtitles provide a concrete textual storyline of the video, \emph{explicitly revealing the underlying logic flows across scenes}.
With these logic flows already exposed in the subtitle stream, VideoSeek can more easily localize answer-critical segments and avoid scanning redundant content, thereby navigating to informative regions with far fewer frame usage but achieving even better performance.
This phenomenon directly supports our earlier assumption that \emph{leveraging the logical flow of videos allows models to use fewer frames while maintaining, or even improving, their video understanding capability}.

\noindent \textbf{Effect of Reasoning Capability}.
This analysis is conducted on LVBench (w/o subtitles).
Table~\ref{tab:think_ablation} reveals two key findings.
First, replacing GPT-5 with GPT-4.1 substantially reduces accuracy from 68.4\% to 53.0\%, while the agent also consumes fewer frames (92.3 vs. 74.2) and performs fewer reasoning turns (4.42 vs. 2.99), indicating that a non-thinking model tends to be over-confident and stops early without sufficient evidence. Second, when GPT-5 is replaced by o4-mini, accuracy drops to 58.5\% despite the agent processing more frames and taking more \emph{think-act-observe} turns, suggesting that reduced reasoning capability damages judgment so that additional computation does not translate into better performance.

\noindent \textbf{Effect of Tool Configurations}.
Table~\ref{tab:tool_ablation} presents the evaluation on LVBench (w/o subtitles). The full toolkit reaches 68.4\%, while removing \texttt{<overview>} causes the largest drop $-13.3$ points, excluding \texttt{<skim>} yields $-6.0$ points, and omitting \texttt{<focus>} gives the smallest drop $-4.7$ points. The performance degradation highlights the importance of all three tools.
\texttt{<overview>} is the most crucial since it provides a global pass over the entire timeline, capturing logic flows throughout the video.

\begin{table}[t]
\centering
\caption{Ablation study on different toolkit configurations. We leave one tool out to validate the significance of earch tool.}\vspace{-2mm}
\begin{adjustbox}{width=0.47\textwidth}\setlength\tabcolsep{11pt}
\begin{tabular}{cccc}
\toprule
\multicolumn{3}{c}{Tool Availability} & LVBench \\
\cmidrule(lr){1-3}
\texttt{<overview>} & \texttt{<skim>} & \texttt{<focus>} & (w/o sub) \\
\midrule
\rowcolor{myblue!15} \cmark & \cmark & \cmark & \textbf{68.4} \\
\midrule
 {\color{red}\xmark} & \cmark & \cmark & \hspace{9.8mm} 55.1 ({\color{red}-13.3}) \\
\cmark &   {\color{red}\xmark} & \cmark & \hspace{8.1mm} 62.4 ({\color{red}-6.0}) \\
\cmark &   \cmark & {\color{red}\xmark} & \hspace{8.1mm} 63.7 ({\color{red}-4.7}) \\
\bottomrule
\end{tabular}\label{tab:tool_ablation}
\end{adjustbox}
\end{table}

\begin{figure*}
    \centering
    \includegraphics[width=0.94\textwidth]{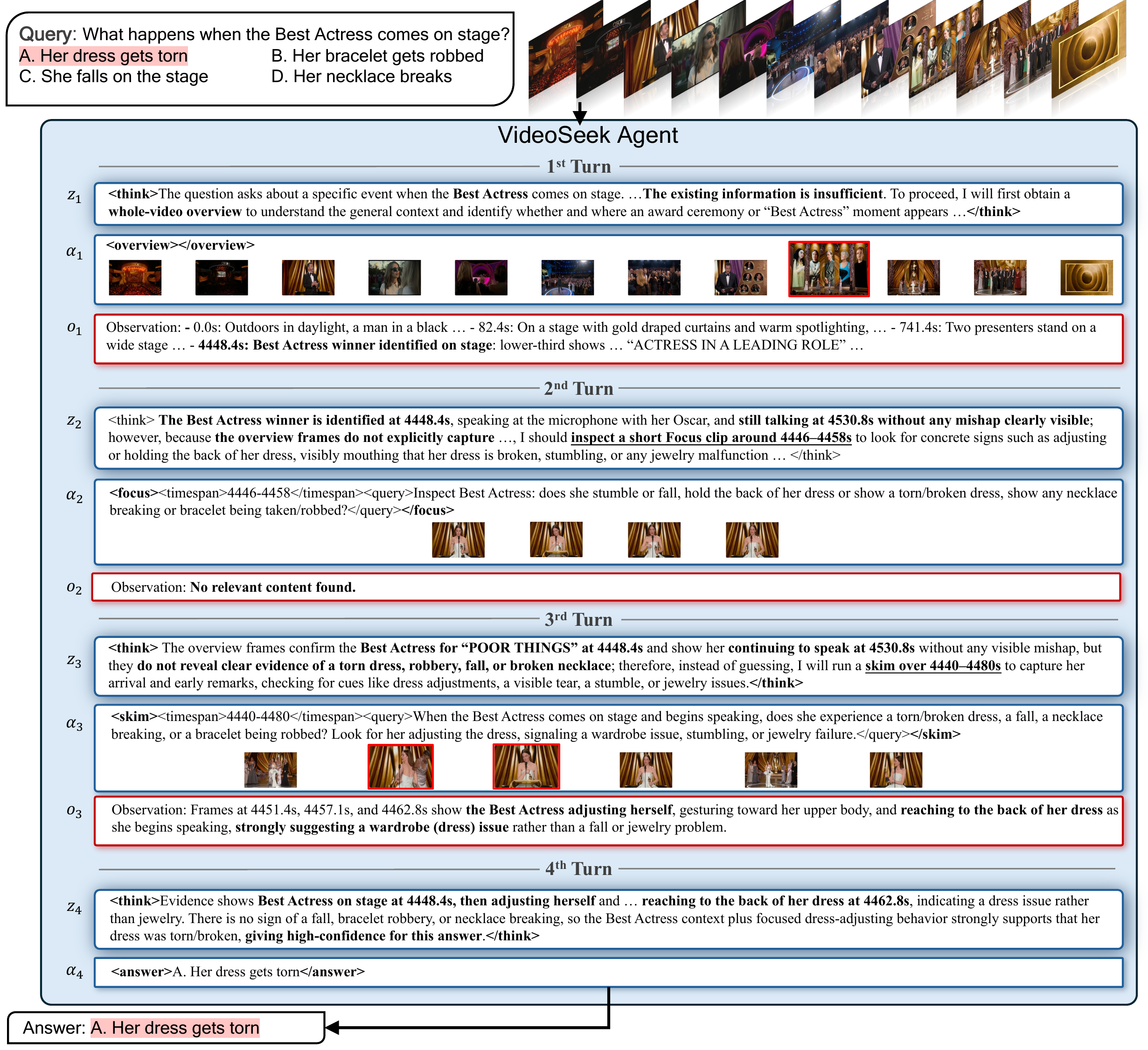}\vspace{-1mm}
    \caption{Case study from LVBench~\cite{wang2025lvbench} (uid: 1671) when applying VideoSeek agent. The example illustrates how the VideoSeek follows a \emph{think–act–observe} loop, reasoning over accumulating observations, then actively invoking \texttt{<overview>}, \texttt{<skim>}, and \texttt{<focus>} tools to inspect only a small subset of frames that are most relevant to the query.}
    \label{fig:case}
\end{figure*}

\noindent \textbf{Case Study}.
Figure~\ref{fig:case} presents a case study that illustrates the long-horizon reasoning process of VideoSeek. In this case, the agent first invokes \texttt{<overview>} to obtain a global storyline of the video and roughly localize potentially relevant moments. It then calls \texttt{<focus>} to inspect a short interval, and finally expands the search with \texttt{<skim>} when the initial evidence is insufficient.
This example illustrates the intended behavior of VideoSeek: progressively refine the search space and gather just enough evidence to answer confidently, rather than densely parsing the full video. Moreover, it also demonstrates that VideoSeek focuses on long-horizon active evidence seeking, where the agent flexibly selects tools to seek useful evidence based on its current state, instead of a predefined ``coarse-to-fine'' rule.

\section{Conclusion}\vspace{-2mm}
We present VideoSeek, a long-horizon video agent that leverages video logic flow to actively seek answer-critical evidence instead of exhaustively parsing the full video. Through a lightweight multi-granular toolkit and a think--act--observe loop, VideoSeek adaptively navigates to informative video segments by reasoning over accumulated observations. Experiments on four challenging benchmarks spanning both long-form video understanding and complex video reasoning show that VideoSeek achieves strong accuracy while using far fewer frames than prior video agents and standalone LMMs. Further analysis highlights the importance of video logic flow, strong reasoning capability, and the complementary design of the toolkit. However, despite its efficiency, VideoSeek may be less suitable for tasks involving unexpected or highly localized surprising moments, such as anomaly detection, where decisive evidence is difficult to anticipate through logic-driven navigation.
Overall, our results suggest that logic-aware, tool-guided seeking is a promising direction for building efficient and scalable video agents, while future work may explore how to better handle rare and unexpected events.
\vspace{-9mm}
{
    \small
    \bibliographystyle{ieeenat_fullname}
    \bibliography{main}
}

\clearpage
\appendix

\section{Appendix}\label{sec:appendix}

This document provides additional empirial analysis and more implementation details of the VideoSeek agent, organized as follows:
\begin{itemize}[labelindent=5pt, labelsep=5pt, leftmargin=*, topsep=0pt]
    \item \textbf{Token Consumption and Runtime} (Section~\ref{sec:token_time}). 
    \item \textbf{Effect of Tool Frame Budget $\alpha$} (Section~\ref{sec:tool_frame}). 
    \item \textbf{Effect of Intermediate Reasoning} (Section~\ref{sec:improve_discuss}).
    \item \textbf{Prompts} (Section~\ref{sec:prompt}).
    \item \textbf{Additional Case Study} (Section~\ref{sec:more_case}).
\end{itemize}

\subsection{Token Consumption and Runtime}\label{sec:token_time}
we report the average frame usage, token consumption, and runtime of GPT-5 and our VideoSeek in Table below:
\begin{table}[h]
\centering
\vspace{-3mm}
\begin{adjustbox}{width=1\linewidth}\setlength\tabcolsep{3pt}
\begin{tabular}{lcccccc}
\toprule[1.2pt]
{Method} & \multicolumn{3}{c}{\bf w/o subtitles} & \multicolumn{3}{c}{\bf w/ subtitles} \\
 \cmidrule(r){2-4} \cmidrule(r){5-7}
\multicolumn{1}{c}{}                        &  \#Frames  & \#Tokens  & {\color{gray} Runtime}  &  \#Frames  & \#Tokens  & {\color{gray} Runtime} \\ \midrule[1.2pt]
GPT-5 (Base) & 384 & 83K & {\color{gray} 66.1s} & 384 & 97K & {\color{gray} 71.3s} \\
VideoSeek & 92.3 & 49K & {\color{gray} 135.9s} &  27.2 & 57K & {\color{gray} 89.7s} \\
\bottomrule[1.2pt]
\end{tabular}
\end{adjustbox}
\vspace{-4mm}
\end{table}

\noindent Compared with the GPT-5 base model, VideoSeek costs substantially fewer frames and fewer tokens, demonstrating that active evidence seeking can improve efficiency. We note that runtime is affected by several hard-to-control factors (\eg, network latency, backend GPU type, API scheduling, and differences in vision/language tokenization and batching). Therefore, we report runtime for completeness, but do not treat it as a fully reliable efficiency metric.

\subsection{Effect of Tool Frame Budget $\alpha$}\label{sec:tool_frame}
Videos on LVBench are substantially longer than those in other benchmarks (LVBench: 67 min; VideoMME-long: 44 min; LongVideoBench-long: 27 min; VideoHolmes: 3 min). To study how the tool frame budget affects the performance–efficiency tradeoff, we define a base configuration controlled by a single scale factor $\alpha$. Specifically, \texttt{<overview>} samples $16\alpha$ frames, \texttt{<skim>} operates on segments of at least $4\alpha$ seconds by uniformly sampling $4\alpha$ frames, and \texttt{<focus>} analyzes clips of at most $4\alpha$ seconds. We then vary $\alpha$ across benchmarks, as shown in Figure below:

\includegraphics[width=0.11\textwidth]{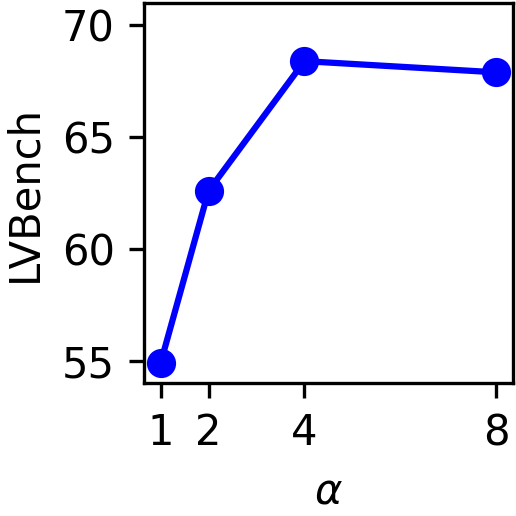}\hfill
\includegraphics[width=0.11\textwidth]{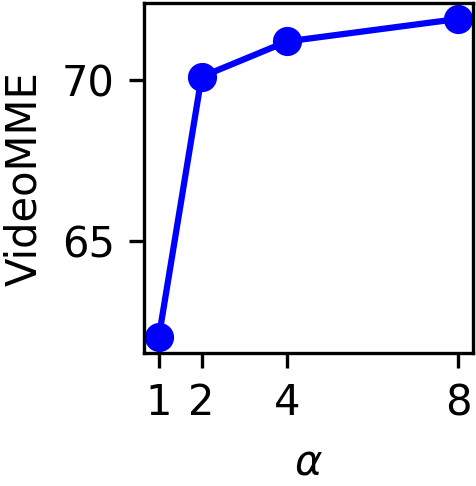}\hfill
\includegraphics[width=0.11\textwidth]{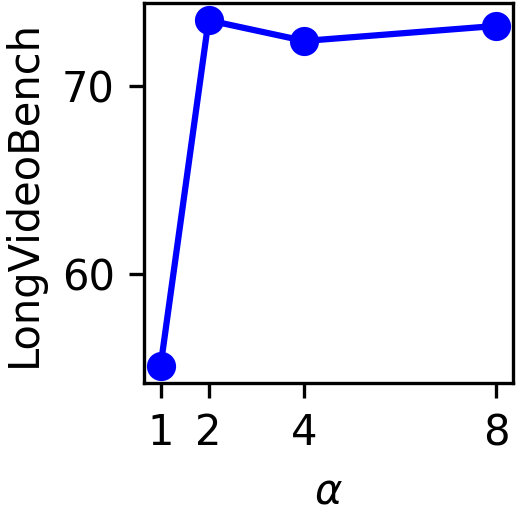}\hfill
\includegraphics[width=0.11\textwidth]{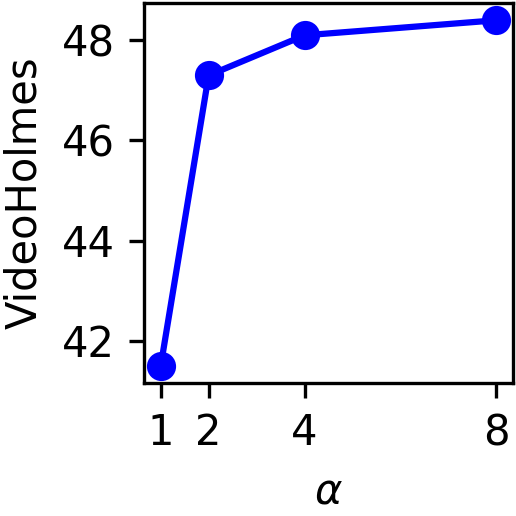}\vspace{-1mm}

\noindent On VideoMME, LongVideoBench, and Video-Holmes, $\alpha = 2$ provides a strong tradeoff between accuracy and efficiency, as most of the gains come from increasing  $\alpha$ from 1 to 2. On LVBench, we observe a similar dominant gain when increasing $\alpha$ from 1 to 4, which is consistent with its substantially longer videos. Based on this trend, we set $\alpha = 4$ for LVBench and $\alpha = 2$ for other three benchmarks.

\subsection{Effect of Intermediate Reasoning}\label{sec:improve_discuss}
We report the results of GPT-5* on LVBench, where GPT-5* is evaluated using the same frames selected by VideoSeek. As shown in Table~\ref{tab:improve_source}, GPT-5* outperforms vanilla GPT-5 while using substantially fewer frames, indicating that VideoSeek’s evidence selection provides more informative visual observations.
Notably, a clear gap remains between GPT-5* and VideoSeek. This gap suggests that the gains of VideoSeek are not solely due to \emph{additional visual evidence}, but also come from the agent’s \emph{intermediate reasoning} over the long-horizon interaction. 
\begin{table}[t]
\centering
\caption{Analysis on effect of intermediate reasoning.}\vspace{-2mm}
\begin{adjustbox}{width=1.0\linewidth}\setlength\tabcolsep{25pt}
\begin{tabular}{lcc}
\toprule
Method & \# Frames & LVBench \\
\midrule
GPT-5 (Base) & 384 & 60.1 \\
GPT-5* & 92.4 & 63.9  \\
VideoSeek & 92.4 & \textbf{68.4} \\
\bottomrule
\end{tabular}
\end{adjustbox}\label{tab:improve_source}
\vspace{-4mm}
\end{table}

\subsection{Prompts}\label{sec:prompt}
\begin{figure*}
    \includegraphics[width=1\textwidth]{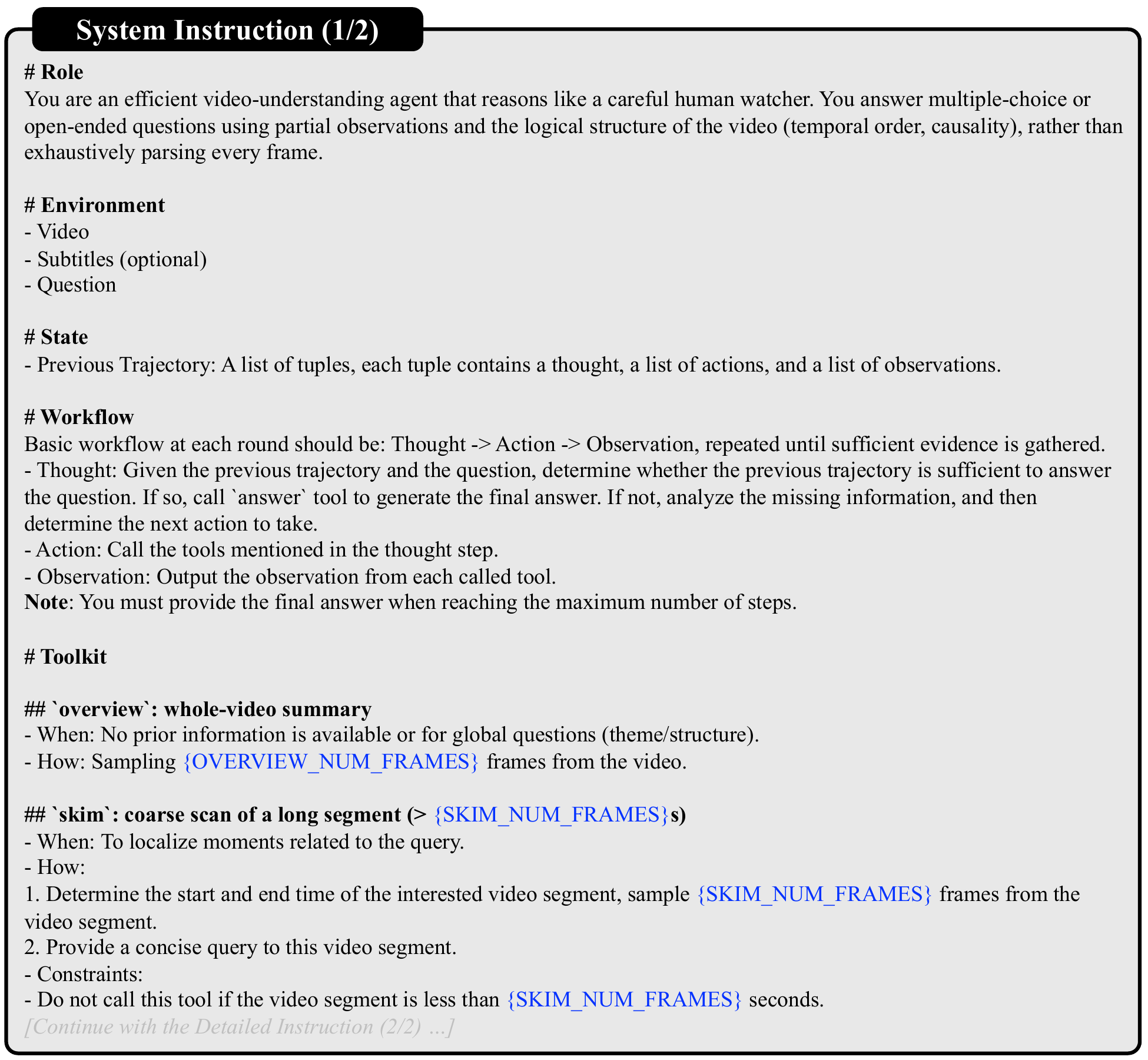}\vspace{0mm}
    \caption{Prompt for the system instruction $\mathcal{I}$ (\emph{part 1}) used in Algorithm 1. {\color{blue} Blue text} denotes variables.}
    \label{fig:sys_instruction1}
\end{figure*}

\begin{figure*}
    \includegraphics[width=1\textwidth]{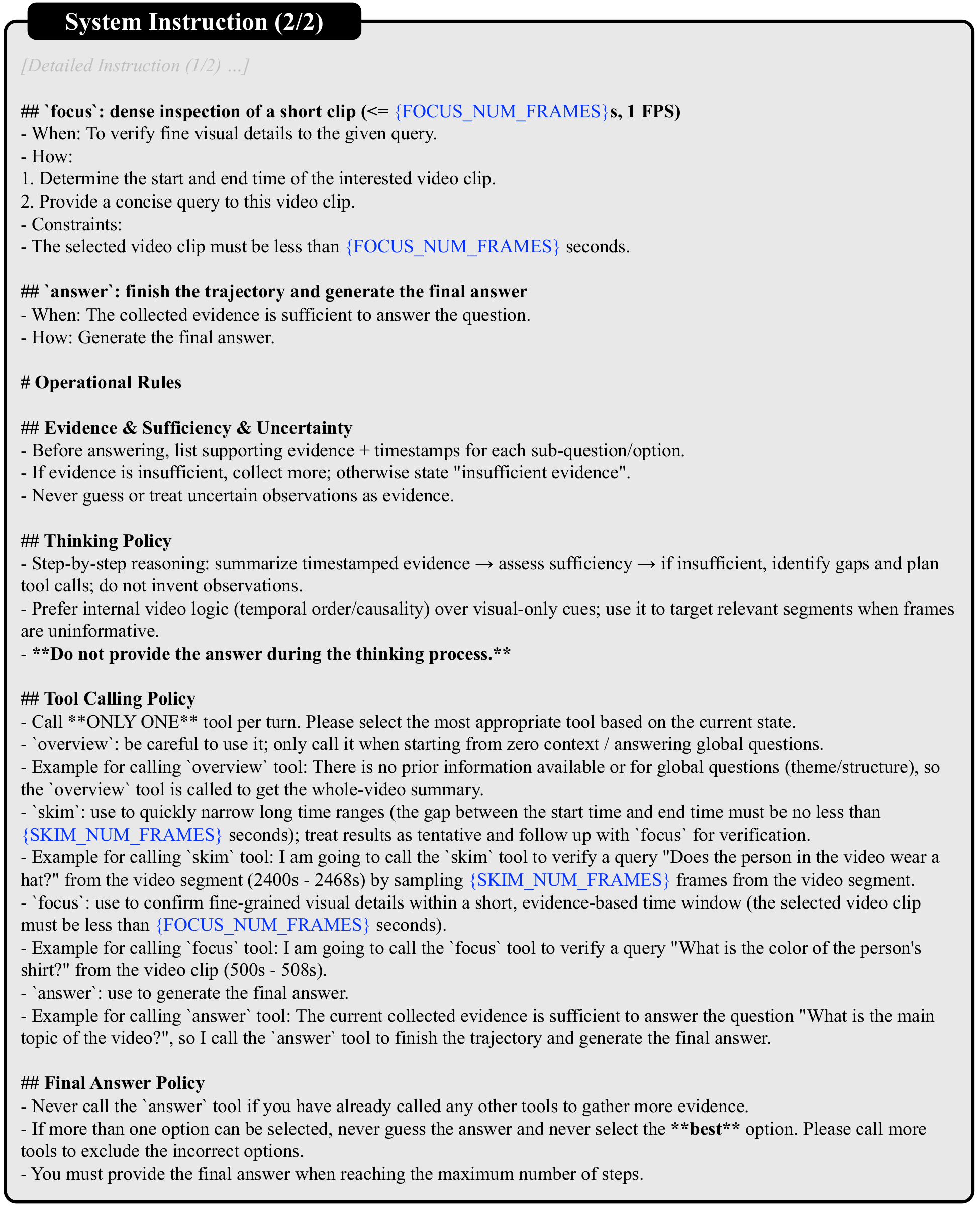}\vspace{0mm}
    \caption{Prompt for the system instruction $\mathcal{I}$ (\emph{part 2}) used in Algorithm 1.}
    \label{fig:sys_instruction2}
\end{figure*}

\noindent \textbf{System Instruction $\mathcal{I}$}. In Figure~\ref{fig:sys_instruction1} and \ref{fig:sys_instruction2}, we provide the exact prompt of the system instruction $\mathcal{I}$ used in Algorithm~\ref{alg:videoseek}.
This instruction is structured into six parts: \textbf{Role}, \textbf{Environment}, \textbf{State}, \textbf{Workflow}, \textbf{Toolkit}, and \textbf{Operational Rules}. Each part defines a different aspect of how the video agent should behave:
\begin{itemize}
    \item \textbf{Role} section specifies that the agent should act as an efficient video-understanding system that reasons like a careful human watcher, answering multiple-choice or open-ended questions from partial observations and the logical structure of the video, rather than exhaustively parsing every frame.
    \item \textbf{Environment} section defines the available inputs to the agent, including the video, optional subtitles, and the question to be answered.
    \item \textbf{State} section specifies the memory available to the agent as a previous trajectory, represented as a list of thought--action--observation tuples that record prior reasoning, tool usage, and collected evidence.
    \item \textbf{Workflow} section describes an iterative \emph{Thought $\rightarrow$ Action $\rightarrow$ Observation} loop, in which the agent first evaluates whether the existing trajectory is sufficient to answer the question, then selects appropriate tools to gather missing evidence, and finally incorporates the resulting observations into subsequent reasoning until sufficient support is obtained or the maximum number of steps is reached.
    \item \textbf{Toolkit} section defines four tools with complementary roles: an \texttt{overview} tool for obtaining a coarse whole-video summary, a \texttt{skim} tool for quickly scanning long video segments to localize potentially relevant moments, a \texttt{focus} tool for densely inspecting short clips to verify fine-grained details, and an \texttt{answer} tool for producing the final response once sufficient evidence has been collected.
    \item \textbf{Operational Rules} section provides practical guidance on how the agent should operate, including collecting timestamped supporting evidence, explicitly checking sufficiency before answering, handling uncertainty without guessing, following disciplined tool-calling constraints, using temporal and causal video logic to guide exploration, and separating intermediate reasoning from the final answer.
\end{itemize}

\begin{figure}
    \includegraphics[width=0.49\textwidth]{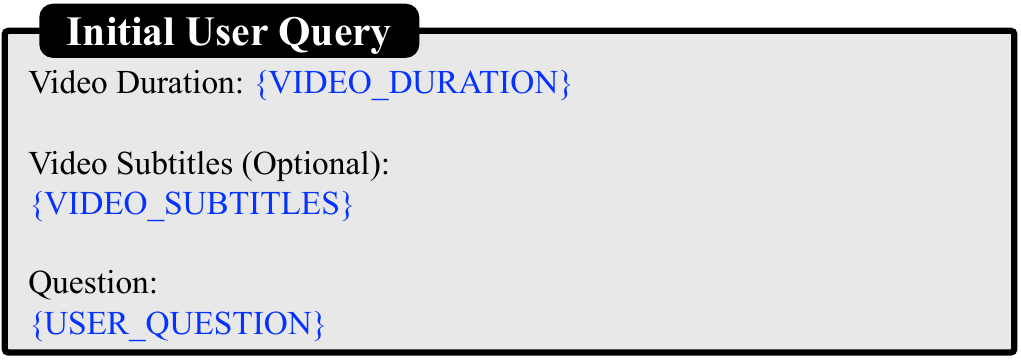}\vspace{0mm}
    \caption{Prompt for the initial user query. {\color{blue} Blue text} denotes variables.}
    \label{fig:init_user_query}
\end{figure}

\noindent \textbf{Initial User Query}. For each Video-QA sample, we first construct an initial user query to trigger the VideoSeek agent’s workflow, as shown in Figure~\ref{fig:init_user_query}. This prompt consists of the video meta information (\ie, video duration and subtitles if available), the user’s question in Figure~\ref{fig:sys_instruction2}.

\begin{figure}
    \includegraphics[width=0.49\textwidth]{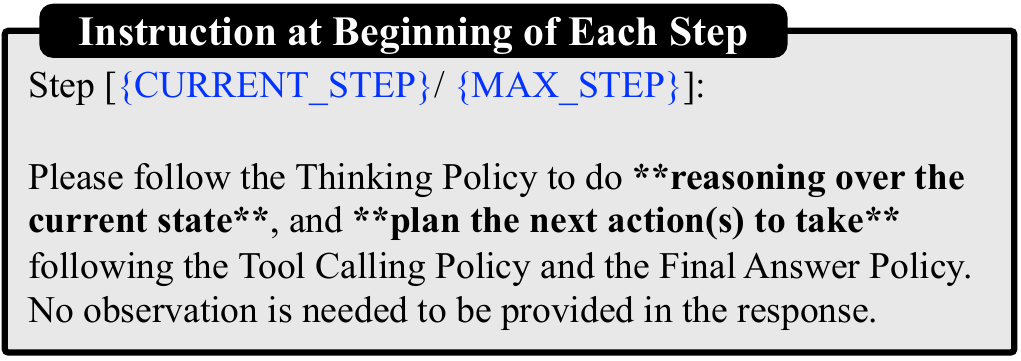}\vspace{0mm}
    \caption{Instruction at the beginning of each step. {\color{blue} Blue text} denotes variables.}
    \label{fig:instruct_begin}
\end{figure}

\noindent \textbf{Instruction at the beginning of each round}. We provide a brief instruction requiring the agent to follow the predefined policies in Figure~\ref{fig:instruct_begin}.

\begin{figure}[t]
    \centering
    \begin{subfigure}{0.49\textwidth}
        \includegraphics[width=\linewidth]{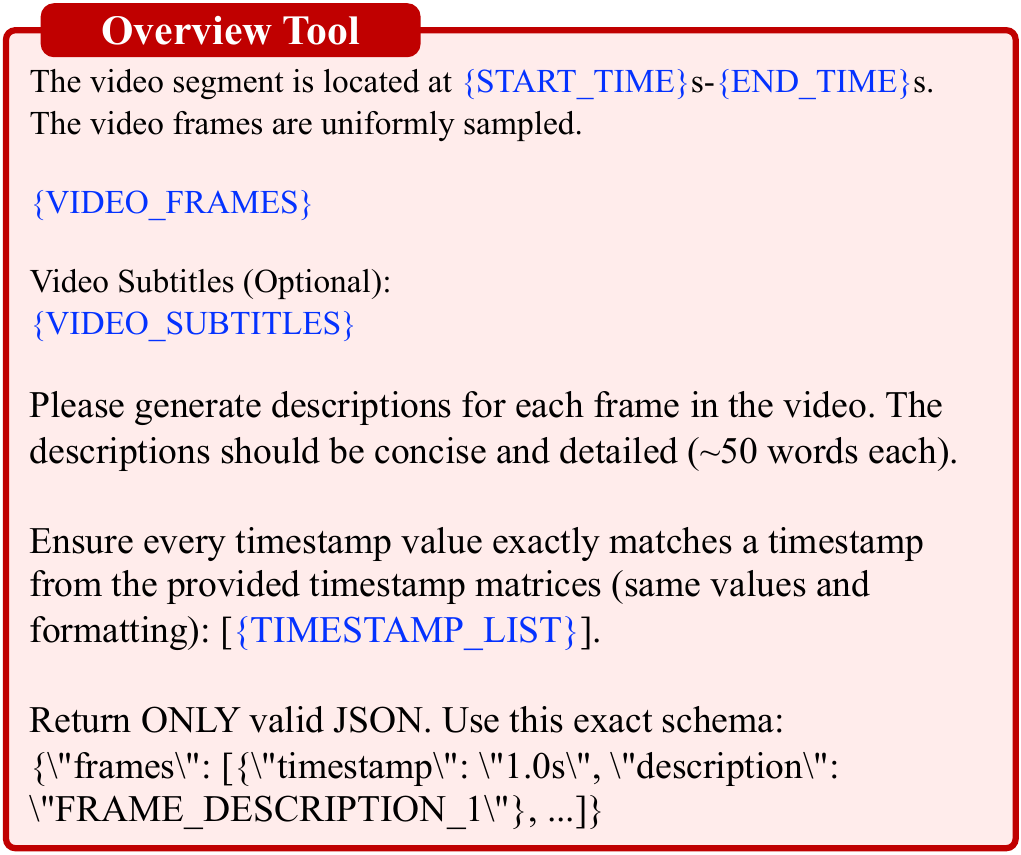}
        \caption{Prompt for calling \texttt{<overview>} tool.}
        \label{fig:overview_tool}\vspace{11mm}
    \end{subfigure}
    \hfill
    \begin{subfigure}{0.49\textwidth}
        \includegraphics[width=\linewidth]{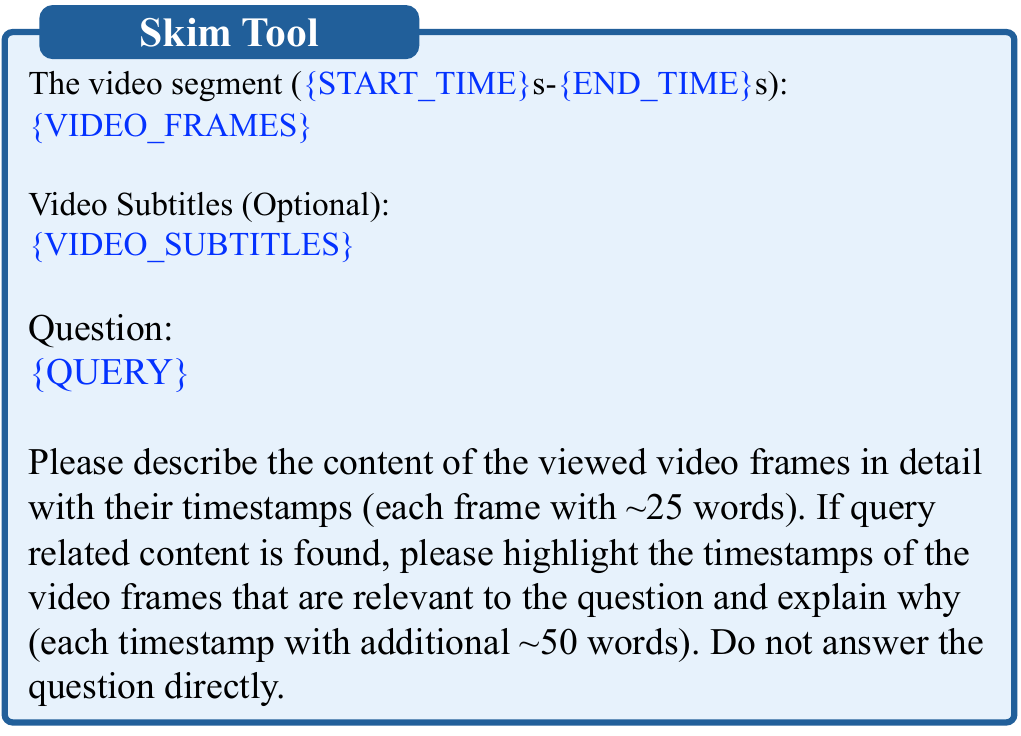}
        \caption{Prompt for calling \texttt{<skim>} tool.}
        \label{fig:skim_tool}\vspace{11mm}
    \end{subfigure}
    \hfill
    \begin{subfigure}{0.49\textwidth}
        \includegraphics[width=\linewidth]{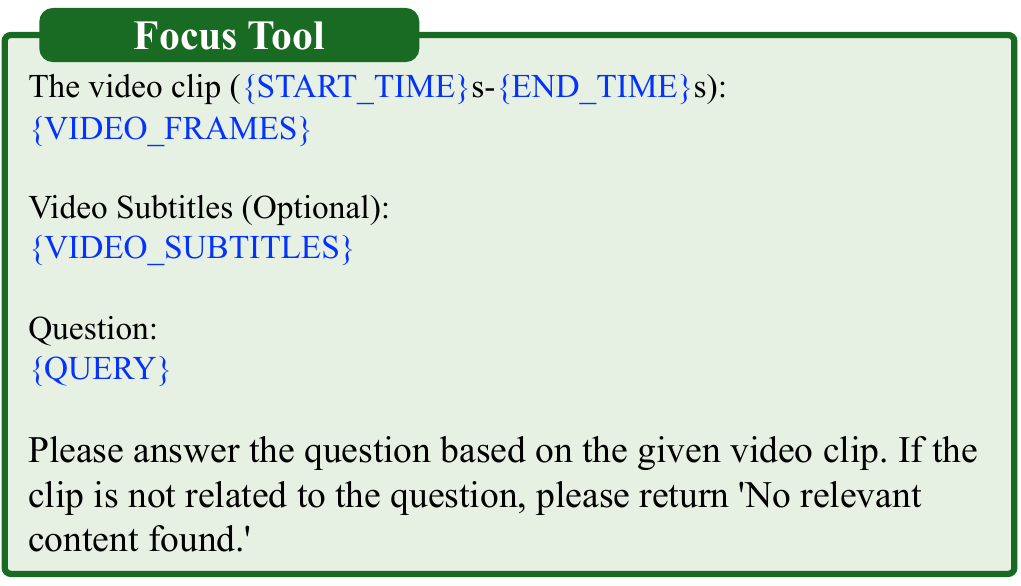}
        \caption{Prompt for calling \texttt{<focus>} tool.}
        \label{fig:focus_tool}
    \end{subfigure}
    \caption{Prompts for tool calling. {\color{blue} Blue text} denotes variables.}
    \label{fig:toolcall_prompt}
\end{figure}

\noindent \textbf{Tool Calling}. The tool-calling prompt are presented in Figure~\ref{fig:toolcall_prompt}. For a given video span, the prompt contains its starting and ending points, the corresponding sampled timestamps, and subtitles (if available), followed by a tool-specific instruction.

\subsection{Additional Case Study}\label{sec:more_case}
\begin{figure*}
    \centering
    \includegraphics[width=1\textwidth]{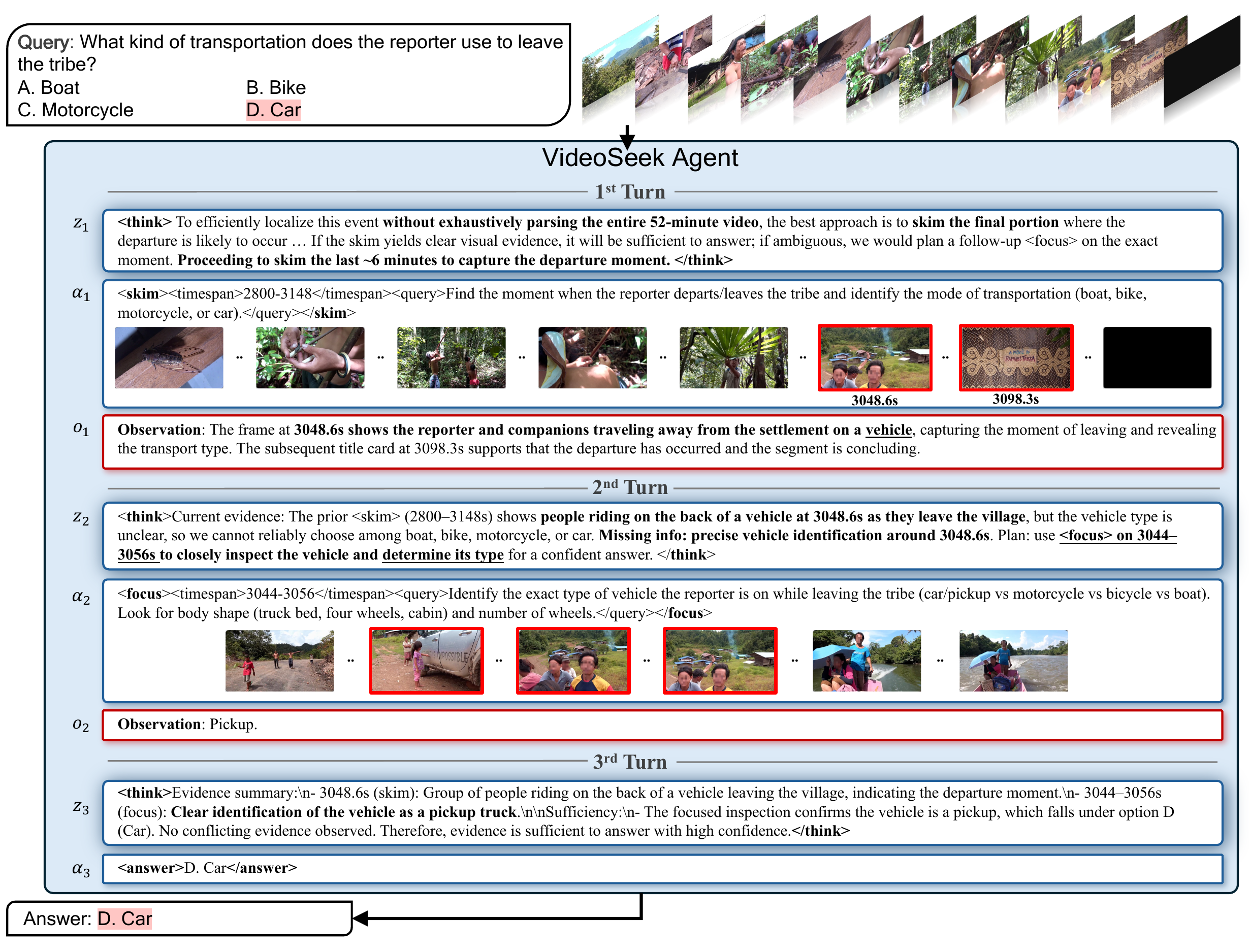}\vspace{-2mm}
    \caption{Case study from LVBench~\cite{wang2025lvbench} (uid: 860) when applying VideoSeek agent.}
    \label{fig:case2}
\end{figure*}

\begin{figure*}
    \centering
    \includegraphics[width=1\textwidth]{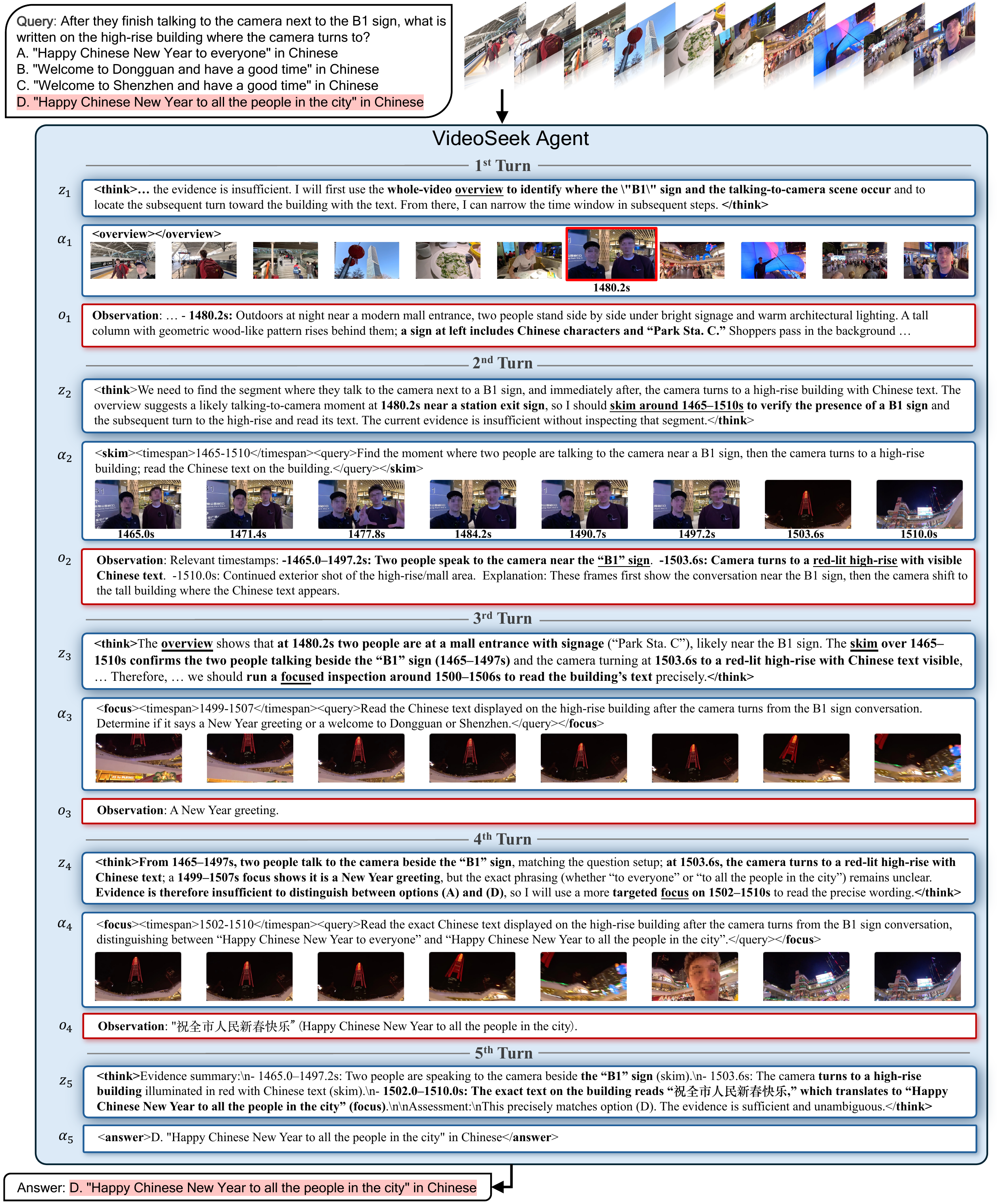}\vspace{-2mm}
    \caption{Case study from LVBench~\cite{wang2025lvbench} (uid: 3105) when applying VideoSeek agent.}
    \label{fig:case3}
\end{figure*}

\begin{figure*}
    \centering
    \includegraphics[width=1\textwidth]{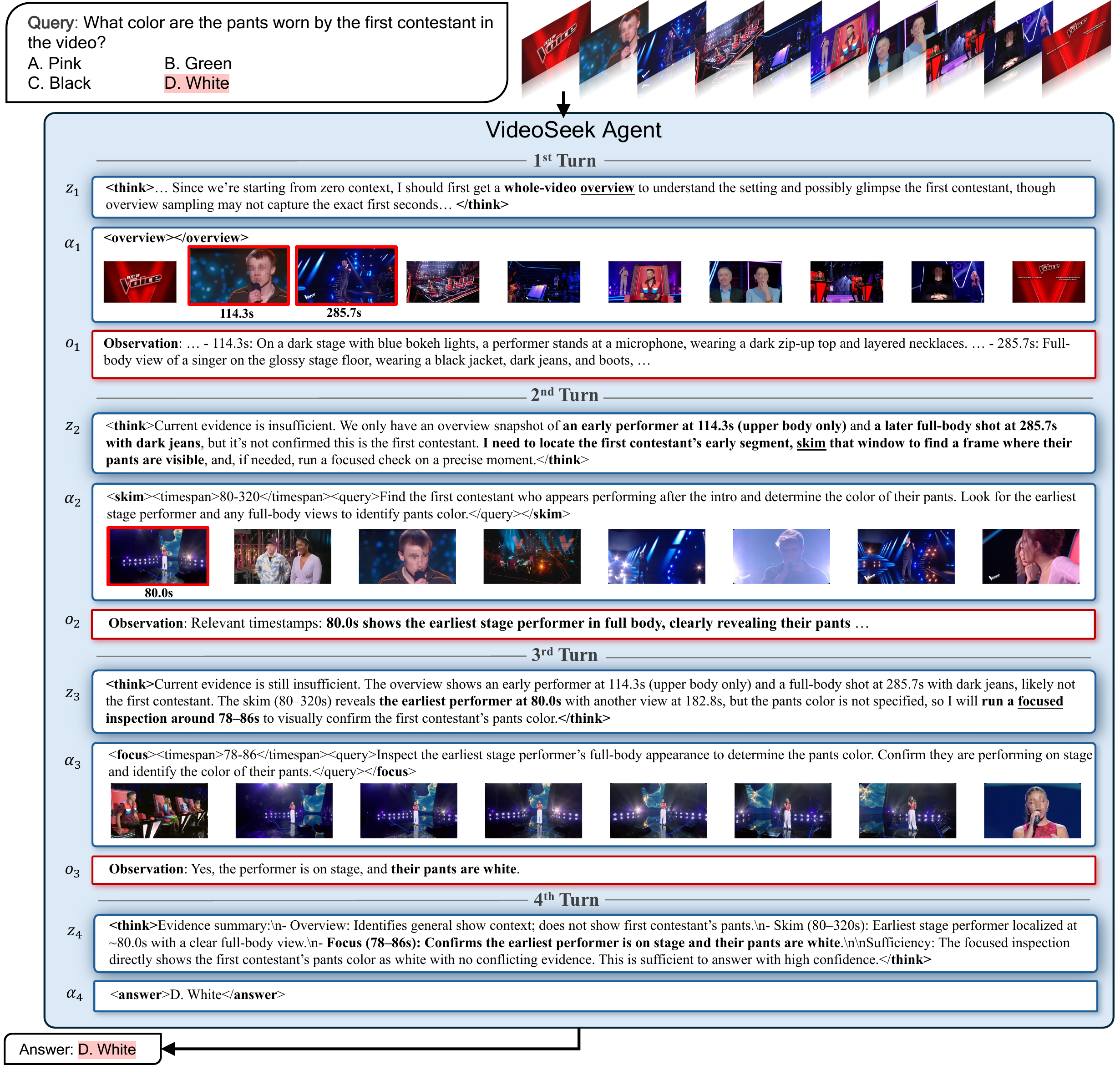}\vspace{-2mm}
    \caption{Case study from LVBench~\cite{wang2025lvbench} (uid: 4490) when applying VideoSeek agent.}
    \label{fig:case4}
\end{figure*}

We present additional case studies showing the representative agentic behavior of VideoSeek, as shown in Figures~\ref{fig:case2}, \ref{fig:case3}, and \ref{fig:case4}.
These examples highlight its key innovation of VideoSeek: \emph{reasoning before observing}, following the video’s logical flow to \emph{actively seek} answer-critical evidence, and executing \emph{long-horizon reasoning} over the accumulating observations.


\end{document}